\pdfoutput=1

\documentclass[11pt]{article}

\usepackage{EACL2023}

\usepackage{times}
\usepackage{latexsym}
\usepackage{subcaption}
\usepackage{enumitem}

\usepackage[T1]{fontenc}

\usepackage[utf8]{inputenc}

\usepackage{microtype}

\usepackage{graphicx}

\title{Vision-Language Models Performing Zero-Shot Tasks Exhibit Gender-based Disparities}

\author{Melissa Hall \qquad Laura Gustafson \qquad Aaron Adcock \qquad Ishan Misra \qquad Candace Ross \\
         Meta AI \\ 
         \\ 
         {\fontsize{9pt}{10pt}\selectfont \texttt{melissahall@meta.com}, \texttt{ccross@meta.com}}}

\usepackage{amsmath}
\usepackage{bbm}
\usepackage{enumitem}
\usepackage{nicefrac}
\usepackage{tikz}
\usetikzlibrary{automata,calc,fit,positioning}
\usepackage{ulem}
\usepackage{wrapfig}

\definecolor{Red}{rgb}{1,0,0}
\definecolor{Green}{rgb}{0,0.69,0}
\definecolor{Blue}{rgb}{0,0,1}
\definecolor{LightBlue}{rgb}{0,0.5,1}
\definecolor{veryLightBlue}{rgb}{0.85,0.98,1}
\definecolor{veryLightGreen}{rgb}{0.6,1,0.6}
\definecolor{Skin}{rgb}{1,0.71,0.69}
\definecolor{Grey}{rgb}{0.5,0.5,0.5}
\definecolor{LightGrey}{rgb}{0.6,0.6,0.6}
\definecolor{VeryLightGrey}{RGB}{219, 219, 219}
\definecolor{Black}{rgb}{0,0,0}
\definecolor{White}{rgb}{1,1,1}
\definecolor{brickred}{rgb}{0.8, 0.25, 0.33}
\definecolor{burntOrange}{RGB}{255,122,20}
\definecolor{navy}{RGB}{80, 74, 255}
\definecolor{teal}{RGB}{0, 123, 159}
\definecolor{aquamarine}{RGB}{51, 153, 255}
\definecolor{saffron}{RGB}{227, 170, 0}
\definecolor{purplePink}{RGB}{160, 89, 107}
\definecolor{xanadu}{RGB}{126, 145, 129}

\newcommand{\white}{\color{White}}

\definecolor{darkgreen}{rgb}{0.0, 0.5, 0.0}
\definecolor{royalblue}{RGB}{50, 92, 168}
\definecolor{coral}{rgb}{1.0, 0.5, 0.31}

\newcommand{\gender}{\texttt{perceived-gender}\ }

\newcommand{\eg}{\textit{e.g.}\ }

\tikzstyle{flow-node}=[draw,ultra thick,minimum height=2.5em,minimum width=2.5em,align=center,rounded corners,fill=LightBlue,font=\centering\white]
\tikzstyle{flow-sub-node}=[draw,dashed,ultra thick,minimum height=2.5em,minimum width=2.5em,align=center,rounded corners,fill=LightBlue,font=\centering\white]
\usepackage{cleveref}

\begin{document}
\maketitle

\begin{abstract}
We explore the extent to which zero-shot vision-language models exhibit gender bias for different vision tasks.
Vision models traditionally required task-specific labels for representing concepts, as well as finetuning; zero-shot models like CLIP instead perform tasks with an open-vocabulary, meaning they do not need a fixed set of labels, by using text embeddings to represent concepts.
With these capabilities in mind, we ask: Do vision-language models exhibit gender bias when performing zero-shot image classification, object detection and semantic segmentation?
We evaluate different vision-language models with multiple datasets across a set of concepts and find (i) all models evaluated show distinct performance differences based on the perceived gender of the person co-occurring with a given concept in the image and that aggregating analyses over all concepts can mask these concerns; (ii) model calibration (i.e. the relationship between accuracy and confidence) also differs distinctly by perceived gender, even when evaluating on similar representations of concepts; and (iii) these observed disparities align with existing gender biases in word embeddings from language models.
These findings suggest that, while language greatly expands the capability of vision tasks, it can also contribute to social biases in zero-shot vision settings. 
Furthermore, biases can further propagate when foundational models like CLIP are used by other models to enable zero-shot capabilities.
\end{abstract}

\section{Introduction}

Natural language has greatly expanded the capabilities of vision models during inference, going from fixed vocabularies of visual concepts to essentially limitless concepts. Vision-language models, such as CLIP~\cite{radford2021learning} and ALIGN \citep{jia2021scaling}, are a powerful means for representation learning of concepts.
These models have impressive zero-shot image recognition capabilities wherein, at test time, the language embeddings corresponding to new visual classes can serve as a classifier.
While such a broad range of recognition abilities is convenient, it also makes these models harder to analyze from a fairness perspective as the model's recognition vocabulary is not fixed and is infinitely large.

Prior works have focused on measuring biases in multimodal word embeddings \citep{ross2021measuring,wang2021assessing} as well as for language and vision models seperately  (\citet{caliskan2017weat,may2019seat,blodgett2020language,fabbrizz2022survey}). 
Other works have measured differential performance of multi-modal models for a small specific vocabulary and find that models contain social biases perpetuating harmful stereotypes.
Examples include disproportionate associations between demographic groups around criminality and dehumanization \citep{agarwal2021evaluating}, with methods using adversarial tuning \cite{https://doi.org/10.48550/arxiv.2202.07603}.
Thus, while works extensively study models with a fixed vocabulary and set of tasks, they do not inspect the performance of language-vision models
in zero-shot settings with an open vocabulary, and they do not investigate how biases may propagate between models.

We measure and explore the gender bias in zero-shot, multi-label image classification by probing CLIP and in the two downstream tasks of object detection using Detic  ~\cite{zhou2022detecting} and semantic segmentation using LSeg ~\cite{li2022languagedriven}.
Probing these three zero-shot vision-language models for gender bias, our contributions are as follows:
\begin{itemize}[noitemsep,nolistsep]
    \item[1.] We show that zero-shot vision-language models show \textit{gender-based performance disparities} for many visual concepts.
    For a given concept, the models perform better when the concept co-occurs with a person with a specific perceived gender as opposed to other genders. 
    Furthermore, we find that aggregating disparity analyses over all concepts can mask these concerns.
    \item[2.] For object detection and segmentation models, the \textit{calibration between model performance and confidence also differs by gender across concepts}. 
    The models treat comparable samples differently depending on the perceived gender of the person in the image.
    \item[3.] Lastly, we find that the \textit{biases in word embeddings}
        \textit{parallel the biases}
        \textit{in the zero-shot vision-language models}. 
    This supports concerns that biases in existing language models may persist in vision tasks that leverage similar language capabilities.
\end{itemize}

\noindent These findings suggest that using language models as supervision may parallel or magnify the biases in vision models, and that biases can propagate when CLIP is used by other models to enable zero-shot capabilities.

\section{Framework}\label{sec:framework}
\subsection{Generating Predictions}\label{sec:metric}
Our analysis focuses on evaluating zero-shot classification, detection, and segmentation models during inference.
In particular, we investigate CLIP$_{ViT-B/32}$ ~\cite{radford2021learning,dosovitskiy2020image}, a contrastive model that, given an image and corresponding text, outputs the cosine similarity between the two modalities using separate language and vision encoders. Due to its design, CLIP can perform zero-shot image classification by using an object class as the text input. 
We also study two models that utilize CLIP for zero-shot detection, Detic \cite{zhou2022detecting}, and semantic segmentation, LSeg \cite{li2022languagedriven}. 
We evaluate these models in the zero-shot setting and they are intentionally not trained on the object classes in the evaluation dataset. 

We use a shared process to allow for a side-by-side of comparison of performance disparities between these three models.
Assume a set of images \textit{I} and a set of object classes \textit{C}, where each image has a set of ground-truth object classes present in the image and a single associated gender label\footnote{We use object classes and concepts interchangeably.}.

For a given image $i \in I$,
\begin{itemize}[nolistsep,noitemsep]
\item[--] a (multi-label) image classifier outputs a set of object classes $C_{pred} \subseteq C$ that are predicted to be contained in the image
\item[--] object detection outputs \textit{N} bounding boxes, where each bounding box is associated with a single label $c_{pred} \in C$
\item[--] semantic segmentation produces a single label $c_{pred} \in C$ for every pixel in the image.
\end{itemize}
We use Detic's bounding box object predictions and LSeg's pixel-by-pixel classifications as binary indicators of the model's recognition of the concept in the image.

\begin{figure*}[h!]
    \centering
    \begin{subfigure}{0.49\textwidth}
        \includegraphics[height=1.2\linewidth]{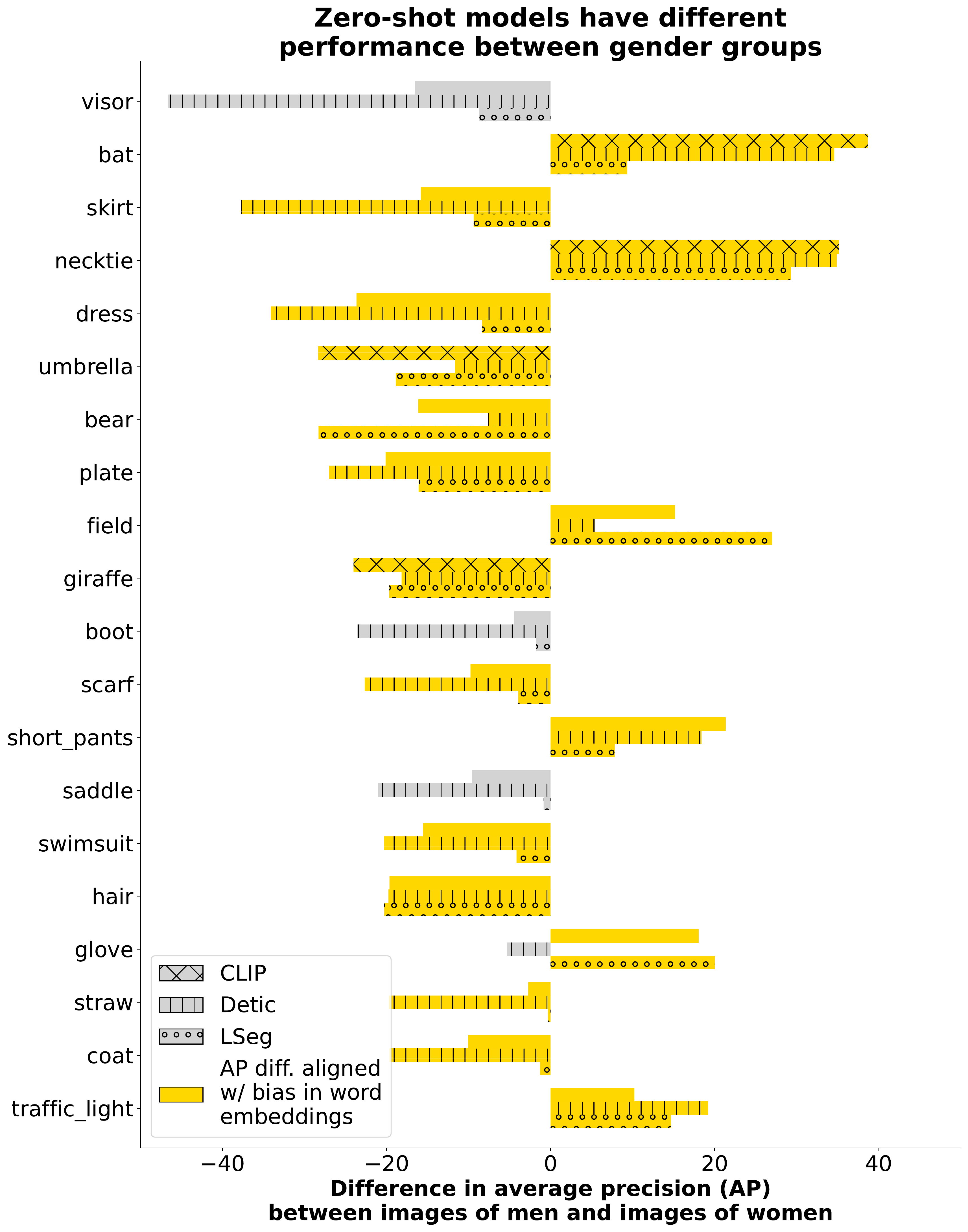}
        \caption{}
        \label{fig:ap_diff}
    \end{subfigure}\hfill
    ~
    \begin{subfigure}{0.49\textwidth}
        \includegraphics[height=1.2\linewidth]{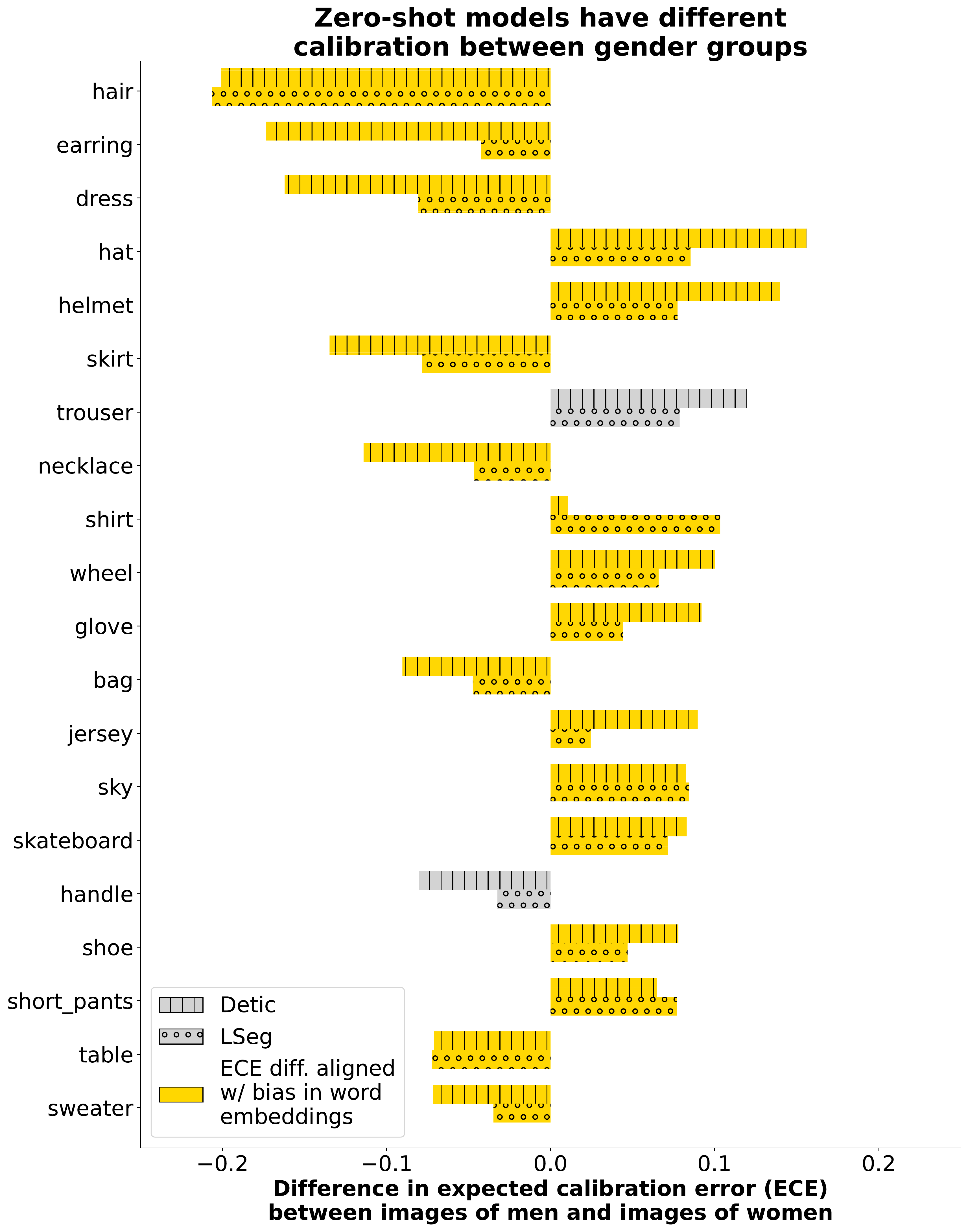}
        \caption{}
        \label{fig:ece_diff}
    \end{subfigure}
    \caption{\textbf{(a)} The average precision (AP) \gender disparity by concept for the Visual Genome, where positive values indicate better performance for images of men, $\mathcal{D}_{men}$, and negative values indicate better performance for images of women, $\mathcal{D}_{women}$. This corresponds to Experiment 1 in Section \ref{sec:exp1}. \textbf{(b)} We measure expected calibration error (ECE), which is the absolute difference between model confidence and accuracy, for each concept. This corresponds to Experiment 2 in Section \ref{sec:exp2}. \textbf{(both)} The bars colored yellow bars show AP and ECE disparities that also align with the social biases measured in word2vec embeddings. Biases in the word2vec embeddings corresponds to Experiment 3 in Section \ref{sec:exp3}.}
    \label{fig:differences}
\end{figure*}

\subsection{Computing Gender-based Disparities}\label{sec:disparities}
For each concept $c \in C$, we explore how the models differ during inference when the gender of the people that are depicted in images changes.
Specifically, we focus on images containing men, $\mathcal{D}_{men}$, versus images containing women, $\mathcal{D}_{women}$.

We primarily use the Visual Genome dataset \cite{krishnavisualgenome}, which contains 108k images.
Each image is annotated with a set of bounding boxes and one object label per box.
The labels correspond to synsets from WordNet \citep{miller1995wordnet}, where each synset is a node representing a singular concept in a tree of nodes.
Visual Genome contains human-annotated synsets tagging objects (\eg labels such as \texttt{baseball bat} and \texttt{dress}) and people (\eg labels such as \texttt{bride}, \texttt{person}, \texttt{doctor}).
We use the synset labels to determine the set of objects in the image and gender of the people present.

To designate gender groups, we create two sets of labels  -- one corresponding to WordNet concepts referring to women (\textit{e.g.} \texttt{mother}, \texttt{wife}) and the other corresponding to WordNet concepts referring to men (\textit{e.g.} \texttt{son}, \texttt{groom})\footnote{We note that these words are used in this manner in WordNet and widely in society, though they can be used across genders.}.
See Table \ref{tab:gendered} in the Appendix for the full synset mapping.
Because these labels are collected from external annotators rather than self-reported by the people shown in the images, we refer to them as \gender labels. 
After getting the \gender labels, we retain only the images that are annotated with a single \gender label and assign them to binary \gender groups.
In Section \ref{sec:limitations}, we discuss with greater detail the challenges and limitations of measuring gender-based performance disparities in this way.
 
To increase reliability of our measurements, we use only the WordNet object labels that occur in at least 50 images for each \gender group.
After filtering, we have 25,215 images and 408 object classes for Visual Genome.

In addition to the Visual Genome dataset, for Experiment 1 we also show results for MS-COCO \cite{lin2014microsoft}, a dataset of 123k images containing a set of bounding boxes for 80 object categories and 5 captions per image. We map the gender-related COCO objects to the Visual
Genome synsets in order to compare differential model performance on similar objects between two different datasets. For each COCO object, we find a Visual Genome synset with the same or similar name. When we have the choice between multiple synsets or multiple synset definitions, we use the Visual Genome synset definition that most closely aligns with the object’s representation in COCO based on visual inspection. 

Because MS-COCO does not include synsets for images, we use the captions to extract the perceived gender(s) in each image similar to previous work \cite{DBLP:journals/corr/ZhaoWYOC17}.
Following our approach for Visual Genome, we create a list of gender-related terms (see Appendix \ref{app:group_annotations} for full list) and keep only those images with a single \gender reference across the captions.
We use the train set for filtering objects that do not occuer in at least 100 images.
This yields 1412 total images and 76 object classes.

\section{Evaluation Setup \& Findings}

We use the process defined in Section \ref{sec:metric} to evaluate whether three vision-and-language models, CLIP, Detic and LSeg, exhibit disparate gender-based outcomes and treatment between concepts, and to what extent these biases are similar to those from language models.
We use images from the Visual Genome with filtered sets of images of men, $\mathcal{D}_{men}$, and woman, $\mathcal{D}_{women}$, as described in Section \ref{sec:disparities}.
There are approximately two times the number of images in $\mathcal{D}_{men}$ than $\mathcal{D}_{women}$\footnote{We note again that these gender definitions are limited, are tied to annotators' perceptions, and use the narrow definitions of binary gender.
}.
We use publicly available pretrained checkpoints for all models.
Detic also requires a minimum score for a predicted bounding box, which we set as 0.1.

\subsection{Experiment 1: Disparity in outcomes by group}\label{sec:exp1}
\paragraph{\textbf{Setup}}
We ask whether a model has similar performance for image classification for a given concept when evaluating images in $\mathcal{D}_{men}$ and $\mathcal{D}_{women}$.
A fair model should have generally comparable performance across demographic groups.
To determine whether each model has differential performance across \gender groups, we use the difference of average precision (AP) between $\mathcal{D}_{men}$ and $\mathcal{D}_{women}$ for every concept.
The average precision is the weighted mean of model precision across concepts. We use AP as it is a popular metric for vision tasks and it accounts for a variable number of objects between images. 
\paragraph{\textbf{Results}} We find that \textbf{\it all models show differential outcomes by gender},
performing disparately between \gender groups $\mathcal{D}_{men}$ and $\mathcal{D}_{women}$ for many concepts.
Figure \ref{fig:differences}(a) shows the greatest differences in average precision (AP) for synsets that co-occur with man- and woman-annotated images for CLIP,  Detic, and LSeg.
Positive AP differences for a concept indicate better outcomes for images in $\mathcal{D}_{men}$ (\eg the model performs better for images containing men over images containing women for the concept \texttt{necktie}) and negative AP differences signal a better measurement for images in $\mathcal{D}_{women}$.
The direction of the AP differences is consistent among all three models for many objects, indicating shared fairness concerns across all of them.

\begin{figure}[h!]
    \centering
    \includegraphics[width=\linewidth]{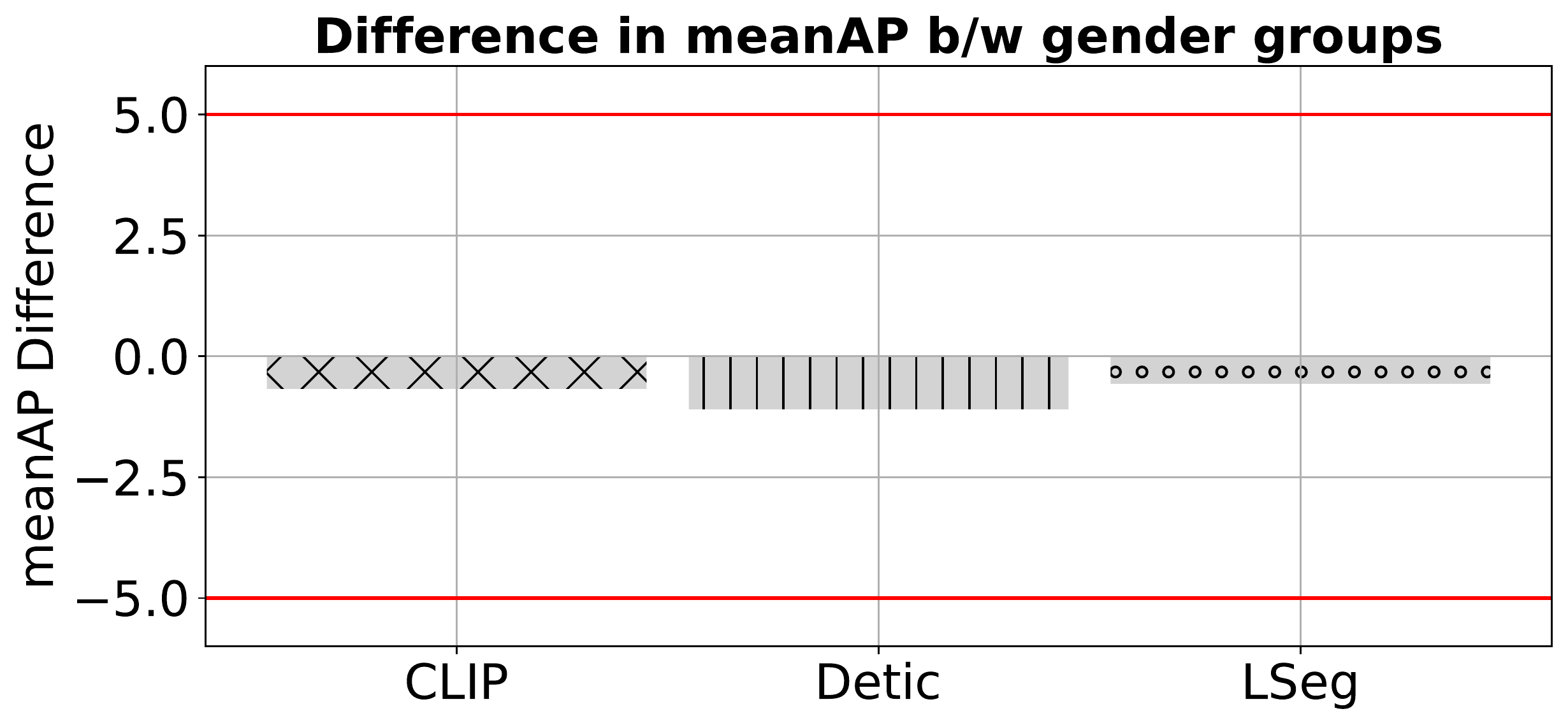}
    \caption{For CLIP, Detic, and LSeg evaluated on Visual Genome, the difference in meanAP between the annotated gender groups masks significant per-concept disparities observed in Figure \ref{fig:differences}.}
    \label{fig:mAP_diff}
\end{figure}

Furthermore, it is common to report the mean average precision across all concepts as a single, summary statistic of model performance (\eg \cite{zhou2022detecting}).
Figure \ref{fig:mAP_diff} demonstrates that \textit{the aggregation of AP across all concepts can mask these disparities} and lend a false assurance of model consistency across demographic groups.

\begin{figure*}[h!]
    \centering
    \begin{subfigure}{0.49\textwidth}
        \includegraphics[height=1.2\linewidth]{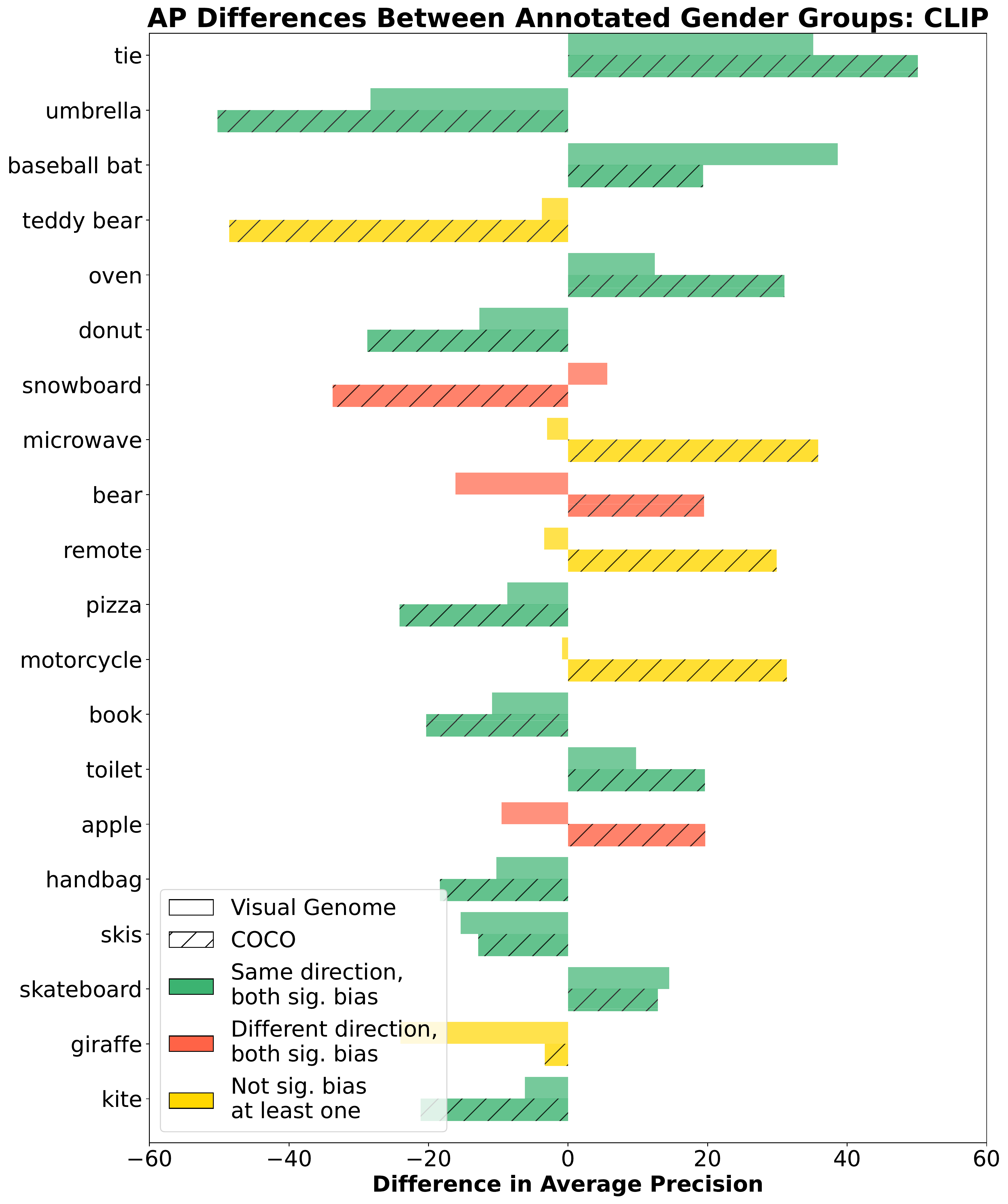}
        \caption{}
        \label{fig:both_datasets_a}
    \end{subfigure}\hfill
    ~
    \begin{subfigure}{0.49\textwidth}
        \includegraphics[height=1.2\linewidth]{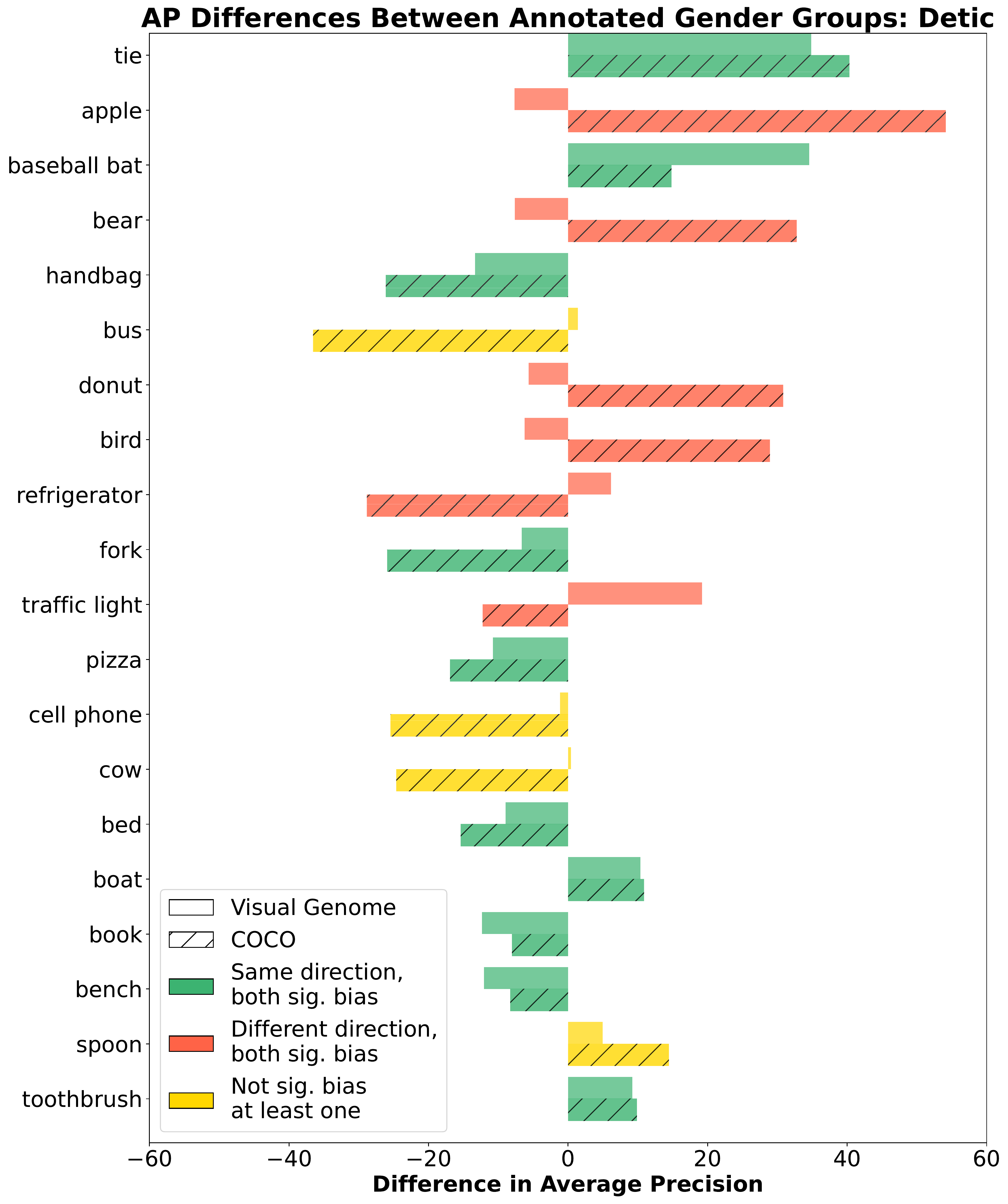}
        \caption{}
        \label{fig:both_datasets_b}
    \end{subfigure}
    \caption{Objects with the highest-disparity in AP for CLIP (\textbf{left}) and Detic (\textbf{right}), evaluated on both Visual Genome and COCO.}
    \label{img:both_datasets}
\end{figure*}

\begin{figure*}[h!]
    \centering
    \begin{subfigure}{\textwidth}
    \hspace{5em}\includegraphics[height=0.25\linewidth]{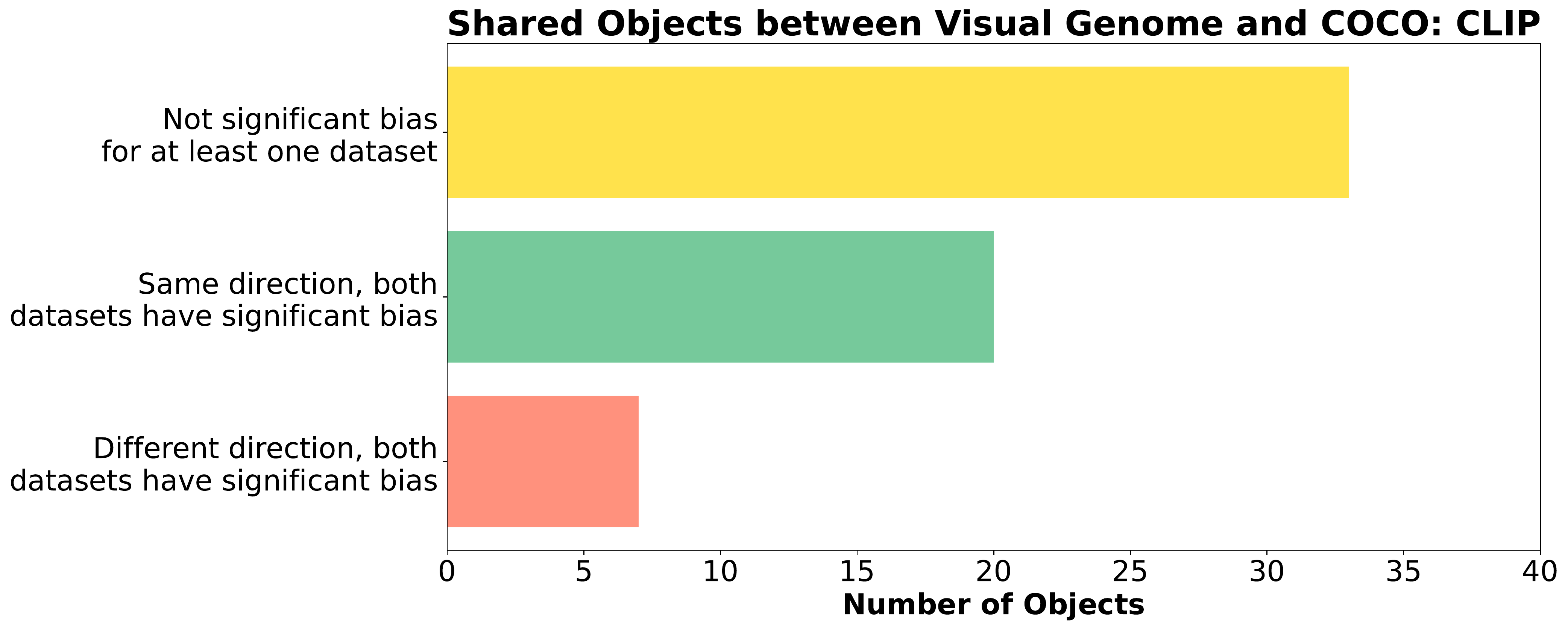}
    \end{subfigure}\hfill

    \begin{subfigure}{\textwidth}
    \hspace{5em}\includegraphics[height=0.25\linewidth]{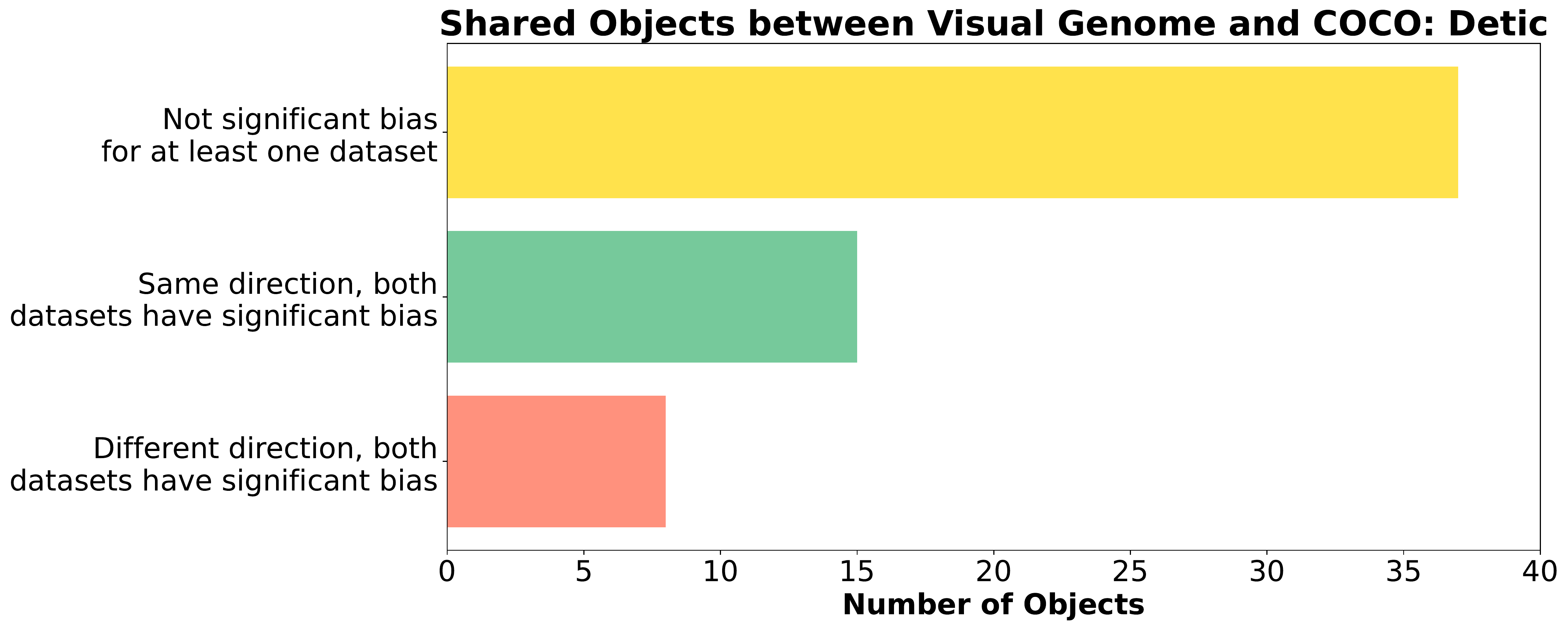}
    \end{subfigure}
    \caption{Many of the objects shared between Visual Genome objects and COCO have a practically significant, directionally similar difference in AP for both CLIP (\textbf{top}) and Detic (\textbf{bottom}).}
    \label{img:both_datasets_all}
\end{figure*}

Lastly, in Figure \ref{img:both_datasets} and Figure \ref{img:both_datasets_all}, we show the overlap between Visual Genome and COCO for a subset of objects in which CLIP and Detic have the highest disparity in AP. To summarize our results, we highlight objects according to whether the AP differences are practically significant and indicate the consistency of disparities between datasets. The color mappings are as follows:
\begin{itemize}[noitemsep,nolistsep]
    \item \textbf{Green}: The AP-differences for both Visual Genome and COCO are in the same direction and have magnitudes greater than 5, meaning that the performance is practically and significantly higher for the same \gender group for both datasets.
    \item \textbf{Red}: The AP-differences for the two datasets are in opposite directions and have magnitudes greater than 5, indicating that the disparity in performance is significant yet inconsistent between the two datasets.
    \item \textbf{Yellow}: At most one AP-difference between the two datasets is greater than 5. This implies that, while the model could favor one group in a dataset over the other, either in the same or different directions, the difference is not practically significant for both datasets.
\end{itemize}
In short, \textit{green} shows alignment in disparities between datasets, \textit{red} shows misalignment, and \textit{yellow} shows either alignment or misalignment but both datasets are not significant.
We omit objects that are not included in both datasets.

We find that there is a significant number of objects with consistent concerns of performance disparity between \gender groups across both datasets for both models (as indicated in green).
Specifically, in Figure \ref{img:both_datasets_all} we see that there are 20 and 15 objects of practically significant, directionally similar performance concerns for both CLIP and Detic respectively, for Visual Genome and COCO.
This provides additional evidence that the concerns observed are not dataset specific and may be pervasive for the given objects even among different distributions of representation.

\subsection{Experiment 2: Disparity in treatment by group}\label{sec:exp2}

\paragraph{Setup} We next explore how models treat different groups using \textit{calibration}. 
If a model's calibration is the similar between groups, it means the model \textit{assigns similar probability scores to samples that have the same expected likelihood of containing the concept}, regardless of which group is depicted in the image. 
For example, suppose we have two images - one containing a man, the other containing a woman - with the same expected likelihood that they contain the concept \texttt{necktie}. 
If the model assigns a lower confidence score to the image with a woman while assigning a higher confidence score to the image with a man, then the model is displaying a disparity in calibration.
Calibration is also closely associated to real-world applications of machine learning systems \cite{https://doi.org/10.48550/arxiv.2103.06172}.

Following previous work \cite{guo2017calibration, DBLP:journals/corr/abs-2201-11706}, we study a model's calibration using the expected calibration error (ECE) \cite{10.5555/2888116.2888120}, which is the absolute difference between the model's confidence and its accuracy.
Larger values of ECE mean greater model miscalibration.
A fair model should have a similar ECE for a given concept regardless of the gender of the person depicted with said concept.
This means ECE$_{\mathcal{D}_{men}} \approx$ ECE$_{\mathcal{D}_{women}}$.
We evaluate Detic and LSeg, excluding CLIP because its multi-label classification setting does not produce probabilities (see Appendix \ref{app:exp2_clip} for further explanation).

\paragraph{Results}
We find that \textit{Detic and LSeg both show differential treatment for gender}.
For each \gender group we're evaluating, we computed the ECE for each concept and took the difference between ECE$_{\mathcal{D}_{men}}$ and ECE$_{\mathcal{D}_{women}}$.
A positive ECE difference means the model is more calibrated for the given concept for images of men and a negative ECE means the model is more calibrated for images of woman.
Figure \ref{fig:differences}(b) shows that many concepts have a large difference in ECE between groups.

\subsection{Experiment 3: Relationship between bias in word embeddings and bias in zero-shot vision-language models}\label{sec:exp3}

\paragraph{Setup} For the final experiment, we explore whether the observed disparities in zero-shot models correlate with disparities found in word embeddings.
Gender bias in word embeddings has been explored using the geometry of the embedding space \citep{bolukbasi2016man} and average cosine similarities between sets of gendered terms and sets of stereotypical, non-gendered terms \citep{caliskan2017weat,may2019seat,tan2019assessing}.
Using a similar approach, we extract embeddings for each concept using word2vec \citep{mikolov2013word2vec} trained on Google News 300M and compute the cosine similarity between each concept and the gendered terms in the word embedding space following the method defined in Appendix \ref{app:cosine_sim}.

\paragraph{Results}

We observe that \textit{the disparities found in Experiments 1 and 2 for zero-shot vision-language models are aligned with those in word embeddings}, as indicated by the cosine similarities.
The bars in Figure \ref{fig:differences} are colored yellow when the difference in cosine similarities between the word2vec text embeddings for the synset and gender terms are aligned with the discrepancies in AP and ECE.
This suggests that language can contribute to propagating social biases in vision-language models, particularly in zero-shot settings that do not perform any further finetuning.

\subsection{Root Cause Analyses}\label{sec:rca}
We analyzed concepts in Visual Genome with disparities across \gender for the three models and found several potential root causes.We see that definitions of concepts like \texttt{necktie} and the salience of objects like \texttt{hair} can differ between groups.
We also find that \textit{how} concepts correlate among each other vary between groups; see Appendix \ref{app:rca}.
These analyses suggest that our findings surface disparities that have real-world implications and can inform potential mitigations.

\section{Discussion}

Natural language supervision has greatly expanded the capabilities of vision models.
Many of these models are able to perform zero-shot image classification, object detection and semantic segmentation on an open-vocabulary.
We probed three models -- CLIP, Detic and LSeg -- to evaluate whether there are gender-based disparities in their performance and treatment for different groups and to see whether these disparities, if any, paralleled those found in word embeddings from language models.

First, we found that these zero-shot models perform differently for many concepts based on the \gender of the co-occurring person in the image across multiple datasets.
We also identified the importance of measuring each concept by itself, as aggregating analyses across concepts can mask the gender-based performance disparities.
Next, we explored the relationship between model confidence and accuracy using the expected calibration error (ECE) and found that the ECE differs for many concepts between \gender groups across all models.
These results show the importance of considering model fairness when using an open vocabulary in zero-shot settings.
Finally, we focused on the relationship between non-contextual word embeddings and the three zero-shot models. The word embeddings have been shown to exhibit gender biases, and these biases are aligned with CLIP, Detic and LSeg.

These findings suggest that using language models as supervision may parallel or magnify the biases in vision models.
Overall, we hope this paves the way for future investigations of these concerns, identifies specific opportunities for mitigations, and serves as an examplar of a method to audit zero-shot vision models for demgraphic disparities.

\section*{Limitations}\label{sec:limitations}
There are multiple issues with using annotations to approximate \gender that plague most image and multi-modal datasets datasets used for disparity evaluations \cite{10.1145/3392866}. 
In many vision-language datasets, captions are primarily crowdsourced or web-scraped, and those depicted in the images are not able to provide their gender identity. 
The captions therefore depict the gender perceptions of those writing the captions, which are limited and rarely include non-binary gender labels.  
The binarization of gender using synsets is reductive and excludes other genders.
Also, this approach relies on annotators’ inherent perception of gender and can lead to the misgendering of individuals depicted.
Reliance on static, external annotations based on visual representations is inherently misaligned with an inclusive operationalization of gender~\cite{devinney2022theories}.
People depicted should be given the agency to optionally share and update their gender information throughout a dataset's lifespan \cite{https://doi.org/10.48550/arxiv.2205.09209}.

Furthermore, while we do our best to perform rigorous and robust measurements, each decision made in a disaggregated evaluation of model performance may affect the observed findings \cite{DBLP:journals/corr/abs-2103-06076}. 
As an example of one such decision, we adapted the detection and segmentation tasks performed by Detic and LSeg as multi-label classification tasks to enable a comparison of the three models between shared metrics and datasets.
Other decisions include the choice of dataset, method of defining groups, and evaluation metric.

\section*{Broader Impacts}
The importance of understanding societal effects of the use of representations from natural language to enable vision tasks only increases as such practices become more ubiquitous.
Model fairness can be defined in many ways and the method of evaluation can reveal different patterns of disparities \cite{DBLP:journals/corr/KleinbergMR16}.
Our study highlights one of several ways to evaluate vision-language systems.

While understanding and minimizing observed disparities in model performance is a valuable goal in itself, it may be insufficient for ensuring that machine learning predictions are unbiased and fair. 
Optimizing models to reduce these disparities requires tradeoffs between other fairness guarantees and performance measures. 

\section*{Acknowledgments}
We thank Adina Williams, Laurens van der Maaten, Nicolas Usunier, Hervé Jegou, Bobbie Chern, Rebecca Qian, and Ranjay Krishna for their feedback on this work.

\clearpage

\bibliography{anthology,custom}
\bibliographystyle{acl_natbib}

\appendix

\section{Appendix}
\label{sec:appendix}

\subsection{Annotations Used for Group Assignment}
\label{app:group_annotations}
The mapping of synset and caption terms used for defining groups in the Visual Genome and COCO datasets are shown in Table \ref{tab:gendered}.

\begin{table*}[h!]
\centering
\begin{subfigure}{0.45\linewidth}
    \centering \textbf{Visual Genome}\\[2ex]
    \begin{tabular}{p{0.4\linewidth} | p{0.4\linewidth}} 
     
     \parbox{\linewidth}{\centering\textit{man-related\\terms}\\[0.8ex]}
     & \parbox{\linewidth}{\centering\textit{woman-related terms}\\[0.8ex]} \\ [0.5ex]
     
     \hline
      man.n.01,\newline male\_child.n.01,\newline guy.n.01,\newline male.n.01,\newline groom.n.01,\newline husband.n.01,\newline grandfather.n.01,\newline son.n.01,\newline boyfriend.n.01,\newline brother.n.01,\newline grandson.n.01,\newline groomsman.n.01,\newline ex-husband.n.01,\newline uncle.n.01,\newline godfather.n.01
      &
      maid.n.02,\newline woman.n.01,\newline girl.n.01,\newline lady.n.01,\newline female.n.01,\newline mother.n.01,\newline lass.n.01,\newline ma.n.01,\newline widow.n.01,\newline bride.n.01,\newline daughter.n.01,\newline grandma.n.01,\newline granddaughter.n.01,\newline bridesmaid.n.01,\newline girlfriend.n.01,\newline sister.n.01,\newline wife.n.01,\newline female\_child.n.01,\newline white\_woman.n.01,\newline dame.n.01,\newline matriarch.n.01,\newline mother\_figure.n.01,\newline dame.n.02,\newline great-aunt.n.01,\newline donna.n.01
      \\
     \end{tabular}
     \end{subfigure}
     ~
     \begin{subfigure}{0.45\linewidth}
         \centering
        \textbf{MS-COCO}\\[2ex]
        \begin{tabular}{p{0.4\linewidth} | p{0.4\linewidth}} 
        
         \parbox{\linewidth}{\centering\textit{man-related\\terms}\\[0.8ex]} &
         \parbox{\linewidth}{\centering\textit{woman-related terms}\\[0.8ex]} \\
         
         \hline
         man,\newline mans,\newline men,\newline boy,\newline boys,\newline father,\newline fathers,\newline son,\newline sons,\newline he,\newline his,\newline him

         & woman,\newline womans,\newline women,\newline girl,\newline girls,\newline lady,\newline ladies,\newline mother,\newline mothers,\newline daughter,\newline daughters,\newline she,\newline her,\newline hers 
         \end{tabular}
        \vspace*{-6.2ex}
     \end{subfigure}

 \caption{The synsets for Visual Genome (left) and words from captions for MS-COCO (right) that we use to determine group membership for gender for the images in the datasets. We exclude the images annotated with synsets or captions that correspond the concept \texttt{person} or \texttt{people} as well as images that correspond to both man-related and woman-related terms.}
\label{tab:gendered}
 \end{table*}

\subsection{Why Experiment 2 Does Not Include CLIP}
\label{app:exp2_clip}
For Experiment 2 as described in Section \ref{sec:exp2}, we measure the disparity in the expected calibration error (ECE) for a given concept across the genders we are evaluating.
We only measure and report ECE for Detic and LSeg.
If we were in a single-label classification setting with \textit{N} object classes, we could take the output of CLIP (cosine similarities ranging from -1 to 1 between the \textit{N} classes and the image) and take the softmax to produce probabilities.
The probabilities correlate to model confidence and can be used to measure the ECE (where both confidences and accuracies range from 0 to 1).
In the multi-label setting, we instead have the raw logits and cannot compute the softmax because multiple classes can be present in the image.
We will consider different approaches in the future to measure calibration in the multi-label setting for CLIP.

\subsection{Cosine Similarity Measurements}
\label{app:cosine_sim}
To perform cosine similarity measurements in Experiment 3, we first define a set of embeddings corresponding to each gender group following prior work \cite{caliskan2017weat}, where each gender group contains multiple related terms. For example, the ``woman'' group set contains terms including ``female'', ``woman'', ``girl'', etc. 
We then define a mapping of embeddings corresponding to each synset: 
We use the synset itself for all concepts where the synset is a canonical term of reference for that concept and the synset consists of only one word (e.g. ``refrigerator.n.01'').
When the concept consists of two words (e.g. ``electric\_refrigerator.n.01''), we average the embeddings between the two words.
When the synset is ambiguous for that concept and may be confused with other synsets related to the same concept, we select a modifier based on the synset definition (e.g. ``knob.n.02'' becomes ``knob handle'' and ``helmet.n.02'' becomes ``protective helmet''). 
For each \gender group, we average the cosine similarity between the synset embeddings and each gender term. 
We then use the difference in the cosine similarites between the ``man'' and ``woman'' gender groups as an indicator of the social bias for that concept: 
When the difference in average cosine similarity is positive, the concept is more aligned with ``man'' terms than ``woman'' terms and vice versa for negative differences in average cosine similarity.

\subsection{Examples from Root Cause Analysis of Disparities}
\label{app:rca}

In the Visual Genome dataset, we observe images of halter swimsuits (which tie at the neck) labeled as ``necktie.n.01'', in addition to the more common necktie traditionally worn with dress-suits.
Because women tend to wear such swimsuits more than men, a model that recognizes ``necktie'' in the more common sense may perform less well for women.

We also see that images that contain representations of hair and dress can vary in salience between gender groups, as the two groups depicted in the photos tend to co-occur with the concepts differently. 
The variation in salience of a given concept between groups likely affects models' predictive performance for the group.

\end{document}